# Reference Distance Estimator

**Yanpeng Li**
Department of Computer Science
Dalian University Technology
Dalian, China 116024
*liyanpeng.lyp@gmail.com*

## Abstract

**Abstract:** A theoretical study is presented for a simple linear classifier called reference distance estimator (RDE), which assigns the weight of each feature $j$ as $P(r|j)-P(r)$, where $r$ is a reference feature relevant to the target class $y$. The analysis shows that if $r$ performs better than random guess in predicting $y$ and is conditionally independent with each feature $j$, the RDE will have the same classification performance as that from $P(y|j)-P(y)$, a classifier trained with the gold standard $y$. Since the estimation of $P(r|j)-P(r)$ does not require labeled data, under the assumption above, RDE trained with a large number of unlabeled examples would be close to that trained with infinite labeled examples. For the case the assumption does not hold, we theoretically analyze the factors that influence the closeness of the RDE to the perfect one under the assumption, and present an algorithm to select reference features and combine multiple RDEs from different reference features using both labeled and unlabeled data. The experimental results on 10 text classification tasks show that the semi-supervised learning method improves supervised methods using 5,000 labeled examples and 13 million unlabeled ones, and in many tasks, its performance is even close to a classifier trained with 13 million labeled examples. In addition, the bounds in the theorems provide good estimation of the classification performance and can be useful for new algorithm design.

## 1  Introduction

Semi-supervised learning [1][2] trains classifiers using both labeled and unlabeled data in order to enhance the classifiers from labeled data only. In many real-world applications, there are much more unlabeled examples available than labeled ones, so the potential development space of semi-supervised learning is large. Various algorithms have been investigated during the past 20 years, such as self-training [3], co-training [4], TSVM [5], EM algorithm [6] and graph-based regularization [7]. Although there are successful applications reported, there is still big challenge in this area [1][2]. For example, it is difficult for large-scale semi-supervised learning to work on the current "big data", and there is little theory that can drive new algorithm design.

Why do people study semi-supervised learning? One intuition is that they hope to move towards a perfect upper bound where all the unlabeled examples are correctly labeled, but it is usually an unreachable goal in practice. We call this model semi-perfect classifier, which is trained by all the gold standard labels in both labeled and unlabeled data. Although it may not be a perfect classifier with 100% accuracy due to the impact of feature engineering, classification algorithm and the nature of data, it can be viewed as an ultimate goal for semi-supervised learning, because if all the unlabeled data is correctly labeled, usually there is no need of semi-supervised learning. Therefore, the theoretical analysis of the distance between certain learning algorithm and its semi-perfect classifier would be important, since from it we can see the potential space for the current method to be further improved.

Consider a simple linear classifier, which assigns the weight of each feature $j$ as $P(r|j)- P(r)$, where $r$ is a Boolean feature. We call this classifier reference distance estimator (RDE) and the feature $r$ reference feature. Consider three different types of RDEs: 1) if $r$ is the gold standard label $y$ in labeled data, it can be viewed as a supervised classifier trained on the labeled data; 2) when $r$ is the gold standard of all the labeled and unlabeled data, it is a semi-perfect classifier described above; 3) once $r$ is not the gold standard label but a feature learned from both labeled and unlabeled data, the RDE works in a semi-supervised setting. In the following sections, we show in theory the impact of some characteristics of the reference features on the distance between a semi-supervised RDE (Case 3) and a semi-perfect RDE (Case 2). Based on the theoretical result, we design an algorithm for constructing good reference features and combining multiple RDEs so as to improve the prediction performance. Experiments of 10 text classification tasks are designed to examine the theory as well as the algorithm based on RDE.

## 2    Reference Distance Estimator

Let the input data be the matrix $\mathbf{X} = [x_{ij}]_{m \times n}$, where $x_{ij} \in \{0,1\}$ indicates the occurrence of feature $j$ in the example $i$ ($x_{ij} = 1$ for "Yes" and $x_{ij} = 0$ for "No"). The row vector $\mathbf{x_i} = [x_{i1} \ldots x_{in}]$ indicates an example, and the column vector $\mathbf{f_j} = [x_{1j} \ldots x_{mj}]^T$ denotes the occurrence of feature $j$ on all the examples. Let $\mathbf{y} = [y_1 \ldots y_m]^T$ be the gold standard class labels, where $y_i \in \{y, \bar{y}\}$ (only the binary case is considered in this work). We define reference distance estimator as a linear combination of features as follows:

$$f(\mathbf{x_i}, r) = \sum_j (P(r|j) - P(r)) x_{ij} \tag{1}$$

where $r$ is called reference feature, and the same for all the examples; $P(j|r)$ is the probability of the feature $j$ conditioned on $r$, and $P(r)$ is the marginal probability of $r$. The probability difference $P(r|j) - P(r)$ as the weight of feature $j$ measurers the change of $P(r)$ after seeing $j$. If $r$ is highly correlated to the class label $y$, intuitively the weight of feature $j$ indicates the correlation information between $j$ and $y$ to some extent, and the linear combination $f(\mathbf{x}, r)$ measures a type of distance between the example $\mathbf{x}$ and the class label $y$. Since the class label is a special Boolean feature, if $r$ is one of the gold standard labels (e.g., $y$ or $\bar{y}$) over all the examples, RDE becomes a linear classifier trained as if all the examples are correctly labeled, exactly the semi-perfect classifier described in Section 1, denoted by $f(\mathbf{x}, y)$ or $f(\mathbf{x}, \bar{y})$, which is intuitively one of the best RDEs derived from different reference features. Although it is not justified by theory, the experimental results in Section 3 demonstrate that $f(\mathbf{x}, y)$ estimated from 13 million labeled examples outperforms a Naïve Bayes classifier trained with the same data. However, in a real world application, it is almost impossible to get the correct class labels for the all the examples directly, and what will the performance be if $r$ is not the gold standard, and how far is it from the semi-perfect RDE? We are trying to give answers in the following theoretical analysis.

### 2.1    Theoretical Analysis

#### 2.1.1    The case under assumption

Before the introduction of the theorems, we give some new definitions related to the theory. For a feature $j$, *feature imbalance coefficient* is defined as:

$$I(j) = \frac{P(j,y) - \alpha P(j,\bar{y})}{P(j)} \tag{2}$$

where $\alpha = P(y)/P(\bar{y})$, $P(j,y)$ and $P(j,\bar{y})$ are the joint probabilities of $j$ and class labels. The metric $I(j)$ measures the imbalance degree of feature $j$ over the positive ($y$) and negative ($\bar{y}$) classes. Obviously, for any $j$, there is $-\alpha \leq I(j) \leq 1$. In the extreme cases, $I(j) = 0$ indicates a feature cannot distinguish positive and negative classes, which is equivalent to random guess. $I(j)$ equals 1 or $-\alpha$ when $j$ appears only in positive or negative examples respectively, that is, $I(y) = 1$ and $I(\bar{y}) = -\alpha$.

**Theorem 1.** *For a reference feature $r$, if $\forall j, P(j,r|y) = P(j|y) P(r|y)$ and $P(j,r|\bar{y}) = P(j|\bar{y}) P(r|\bar{y})$, then $\forall i$*

$$f(\mathbf{x_i}, r) = \frac{P(r)I(r)}{\alpha} \sum_j I(j) x_{ij} \tag{3}$$

**Proof**: $f(\mathbf{x_i}, r) = \sum_j (P(r|j) - P(r)) x_{ij}$ (from Formula (1))

$= \sum_j \frac{1}{P(j)} (P(r,j|y) P(y) + P(r,j|\bar{y}) P(\bar{y}) - P(j) P(r)) x_{ij}$

Using the assumption $P(j,r|y) = P(j|y) P(r|y)$, and $P(j,r|\bar{y}) = P(j|\bar{y}) P(r|\bar{y})$), we have

$f(\mathbf{x_i}, r) = \sum_j \frac{1}{P(j)} (P(r|y)P(j|y)P(y) + P(r|\bar{y})P(j|\bar{y})P(\bar{y}) - P(j) P(r)) x_{ij}$

$= \frac{1}{P(j)} \sum_j (P(r,y)P(j,y)/P(y) + P(r,\bar{y})P(j,\bar{y})/P(\bar{y}) - (P(j,y) + P(j,\bar{y}))(P(r,y) + P(r,\bar{y}))) x_{ij}$

$= \frac{1}{\alpha P(j)} \sum_j ((P(j,y) - \alpha P(j,\bar{y}))(P(r,y) - \alpha P(r,\bar{y}))) x_{ij}$

$= \frac{P(r)I(r)}{\alpha} \sum_j I(j) x_{ij}$ (according to Formula (2))

The theorem implies that if $r$ is conditionally independent with each feature $j$ on both classes $y$ and $\bar{y}$, the decision function can be written as the product of two parts $P(r)I(r)/\alpha$ and $\sum_j I(j) x_{ij}$, where the impact of reference feature and original features can be described separately. From Theorem 1, we can get an interesting corollary as below:

**Corollary 1.** *For a reference feature $r$, if*

1) $I(r) \neq 0$ *and*

2) $\forall j, P(j,r|y) = P(j|y) P(r|y)$, *and* $P(j,r|\bar{y}) = P(j|\bar{y}) P(r|\bar{y})$, *then*

1) $\forall i, \quad \frac{f(\mathbf{x_i},r)}{P(r)I(r)} = \frac{(1+\alpha)f(\mathbf{x_i},y)}{\alpha} = -\frac{(1+\alpha)f(\mathbf{x_i},\bar{y})}{\alpha} = \frac{1}{\alpha}\sum_j I(j) x_{ij}$ (4)

2) $\forall i$, *the classifier $f(\mathbf{x_i}, r)$ yields the same ROC curve as one of $f(\mathbf{x_i}, y)$ or $f(\mathbf{x_i}, \bar{y})$ on the task of predicting $y_i$.*

**Proof:** For the semi-perfect RDE $f(\mathbf{x}, y)$ we have $I(y) = \frac{P(y,y) - \alpha P(y,\bar{y})}{P(y)} = \frac{P(y) - 0}{P(y)} = 1$,

$P(j, y|y) = P(j|y) = P(j|y) P(y|y)$ and $P(j, y|\bar{y}) = 0 = P(j|\bar{y}) P(y|\bar{y})$ (conditional independence).

Hence using Theorem 1, we have:

$$f(\mathbf{x_i}, y) = \frac{P(y)I(y)}{\alpha} \sum_j I(j) x_{ij} = \frac{1}{(1+\alpha)} \sum_j I(j) x_{ij} \tag{5}$$

Similarly, for $h(\mathbf{x}, \bar{y})$, we have:

$$f(\mathbf{x_i}, \bar{y}) = -\frac{1}{(1+\alpha)} \sum_j I(j) x_{ij} \tag{6}$$

If $I(r) \neq 0$, combining Formula (3), (5) and (6), we have:

$\frac{f(\mathbf{x_i},r)}{P(r)I(r)} = \frac{(1+\alpha)f(\mathbf{x_i},y)}{\alpha} = -\frac{(1+\alpha)f(\mathbf{x_i},\bar{y})}{\alpha} = \frac{1}{\alpha}\sum_j I(j) x_{ij}$. This concludes the proof of the first part.

For the second part, if $I(r) > 0$, $\forall i, k$ if $f(\mathbf{x_i}, r) \geq f(\mathbf{x_k}, r)$, using Formula (4), we have $f(\mathbf{x_i}, y) \geq f(\mathbf{x_k}, y)$, so $f(\mathbf{x}, r)$ and $f(\mathbf{x}, y)$ yield the same ranking order for each example, thus leading to the same ROC curve. Similarly, if $I(r) < 0$, $f(\mathbf{x}, r)$ and $f(\mathbf{x}, \bar{y})$ have the same ROC curve. The proof is completed.

In other words, if $r$ performs better than random guess and conditionally independent with each feature $j$ on both classes, the RDE $f(\mathbf{x}, r)$ will have the same classification performance with one of the semi-perfect RDEs $f(\mathbf{x}, y)$ and $f(\mathbf{x}, \bar{y})$, since given the proper threshold the same ROC curve always means the same result on almost all the evaluation measures for classification such as Precision, Recall, Accuracy, F-score and AUC. Using Formula (4), we have $f(\mathbf{x_i}, \bar{y}) = -f(\mathbf{x_i}, y)$, so the two semi-perfect classifiers tend to yield the same result in practice, since usually completely opposite decision functions can be easily integrated to the same one using training data.

### 2.1.2 The real-world case

In the real-world application, sometimes it is difficult to find a reference feature exactly under the assumption of Corollary 1, and how the performance will be when the assumption does not hold? Since $\frac{f(\mathbf{x},r)}{P(r)I(r)}$ and $\frac{(1+\alpha)f(\mathbf{x},y)}{\alpha}$ are equivalent when the assumption holds, for the real-world case we investigate the distance between them defined as bellow:

$$Dist(f(\mathbf{x},r), f(\mathbf{x},y)) = \sum_i P(i) \left| \frac{f(\mathbf{x_i},r)}{P(r)I(r)} - \frac{(1+\alpha)f(\mathbf{x_i},y)}{\alpha} \right| \qquad (7)$$

It is the expectation of the absolute value of the difference between a RDE and a semi-perfect RDE. Keeping it small tends to make a RDE close to the semi-perfect classifier.

For any feature $j, r,$ and $l$, we define *conditional dependence coefficient* as:

$$D(j,r|l) = \begin{cases} \frac{P(j,r|l)}{P(j|l)P(r|l)} - 1, & P(j|l) \neq 0 \text{ and } P(r|l) \neq 0 \\ 0, & P(j|l) = 0 \text{ or } P(r|l) = 0 \end{cases} \qquad (8)$$

It describes the degree of conditional dependence between two features $j$ and $r$. The two features are conditionally independent when $D(j,r|l) = 0$. This measure was found to play a key role in bounding the distance in Formula (7).

**Lemma 1.** For $\forall r, j$, there is:

$$\frac{P(j,r)}{P(j)P(r)} - 1 - \frac{I(r)I(j)}{\alpha} = \frac{(\alpha+I(j))(\alpha+I(r))}{\alpha(1+\alpha)} D(j,r|y) + \frac{(1-I(j))(1-I(r))}{(1+\alpha)} D(j,r|\bar{y}) \qquad (9)$$

**Proof:** if $P(j|y) \neq 0, P(r|y) \neq 0, P(j|\bar{y}) \neq 0, P(r|\bar{y}) \neq 0$

$$\frac{(\alpha+I(j))(\alpha+I(r))}{\alpha(1+\alpha)} D(j,r|y) + \frac{(1-I(j))(1-I(r))}{(1+\alpha)} D(j,r|\bar{y})$$

$$= \frac{(\alpha+I(j))(\alpha+I(r))}{\alpha(1+\alpha)} \left( \frac{P(j,r|y)}{P(j|y)P(r|y)} - 1 \right) + \frac{(1-I(j))(1-I(r))}{(1+\alpha)} \left( \frac{P(j,r|\bar{y})}{P(j|\bar{y})P(r|\bar{y})} - 1 \right)$$

$$= \frac{(\alpha+I(j))(\alpha+I(r))}{\alpha(1+\alpha)} \frac{P(j,r|y)}{P(j|y)P(r|y)} + \frac{(1-I(j))(1-I(r))}{(1+\alpha)} \frac{P(j,r|\bar{y})}{P(j|\bar{y})P(r|\bar{y})} - 1 - \frac{I(r)I(j)}{\alpha}$$

$$= \frac{(1+\alpha)P(y|j)P(y|r)}{\alpha} \frac{P(j,r|y)}{P(j|y)P(r|y)} + (1+\alpha)P(\bar{y}|j)P(\bar{y}|r) \frac{P(j,r|\bar{y})}{P(j|\bar{y})P(r|\bar{y})} - 1 - \frac{I(r)I(j)}{\alpha}$$

$$= \frac{P(j,r)}{P(j)P(r)} - 1 - \frac{I(r)I(j)}{\alpha}$$

If $P(j|y) = 0$, then $D(j,r|y) = 0$, $P(j,r) = P(j,r,\bar{y})$, $P(j|\bar{y}) = P(j)/P(\bar{y})$, and $I(j) = -\alpha$.

Therefore, $\frac{(\alpha+I(j))(\alpha+I(r))}{\alpha(1+\alpha)} D(j,r|y) + \frac{(1-I(j))(1-I(r))}{(1+\alpha)} D(j,r|\bar{y}) = \frac{(1-I(j))(1-I(r))}{(1+\alpha)} \left( \frac{P(j,r|\bar{y})}{P(j|\bar{y})P(r|\bar{y})} - 1 \right)$

$$= \frac{(1+\alpha)(1-I(r))}{(1+\alpha)} \left( \frac{P(j,r|\bar{y})}{P(j)/P(\bar{y})P(r|\bar{y})} - 1 \right) = \frac{P(j,r,\bar{y})}{P(j)P(r)} - 1 + I(r) = \frac{P(j,r)}{P(j)P(r)} - 1 - \frac{I(r)I(j)}{\alpha}$$

Similarly, we can obtain that if any of $P(j|y), P(r|y), P(j|\bar{y}),$ and $P(r|\bar{y})$ equals zero, Formula (9) will be satisfied. This concludes the proof of Lemma 1.

This lemma implies that the difference between the weights of a RDE and a Semi-perfect RDE (the left side of Formula (9)) can be represented as a linear combination of $D(j,r|y)$ and $D(j,r|\bar{y})$ (the right side of Formula (9)). Based on it we can obtain the relation between the distance in Formula (7) and conditional dependence coefficients described in the following theorems.

**Theorem 2.** *Given a RDE $f(\mathbf{x},r)$ and a semi-perfect RDE $f(\mathbf{x},y)$, we have:*

$$Dist(f(\mathbf{x},r), f(\mathbf{x},y)) \leq \frac{1}{|I(r)|} \sum_j P(j) \left( \frac{(\alpha+I(j))(\alpha+I(r))}{\alpha(1+\alpha)} |D(j,r|y)| + \frac{(1-I(j))(1-I(r))}{(1+\alpha)} |D(j,r|\bar{y})| \right) \qquad (10)$$

$$\leq \frac{1}{|I(r)|}\sum_j P(j) \max\left(\frac{(\alpha+I(r))}{\alpha}|D(j,r|y)|, (1-I(r))|D(j,r|\bar{y})|\right) \quad (11)$$

**Proof:** According to Formula (4) and (7),

$$Dist(f(\mathbf{x},r), f(\mathbf{x},y)) = \frac{1}{|I(r)|}\sum_i P(i)\left|\sum_j \left(\frac{P(j,r)}{P(j)P(r)} - 1\right)x_{ij} - \frac{I(r)}{\alpha}\sum_j I(j)x_{ij}\right|$$

$$= \frac{1}{|I(r)|}\sum_i P(i)\left|\sum_j \left(\frac{P(j,r)}{P(j)P(r)} - 1 - \frac{I(r)I(j)}{\alpha}\right)x_{ij}\right|$$

$$= \frac{1}{|I(r)|}\sum_i P(i)\left|\sum_j \left(\frac{(\alpha+I(j))(\alpha+I(r))}{\alpha(1+\alpha)}D(j,r|y) + \frac{(1-I(j))(1-I(r))}{\alpha}D(j,r|\bar{y})\right)x_{ij}\right| \quad \text{(Lemma 1)}$$

$$\leq \frac{1}{|I(r)|}\sum_i P(i)\sum_j\left(\left|\frac{(\alpha+I(j))(\alpha+I(r))}{\alpha(1+\alpha)}D(j,r|y)\right| + \left|\frac{(1-I(j))(1-I(r))}{\alpha}D(j,r|\bar{y})\right|\right)x_{ij}$$

$$= \frac{1}{|I(r)|}\sum_i P(i)\sum_j\left(\frac{(\alpha+I(j))(\alpha+I(r))}{\alpha(1+\alpha)}|D(j,r|y)| + \frac{(1-I(j))(1-I(r))}{\alpha}|D(j,r|\bar{y})|\right)x_{ij}$$

$$= \frac{1}{|I(r)|}\sum_j\left(\sum_i P(i)x_{ij}\right)\left(\frac{(\alpha+I(j))(\alpha+I(r))}{\alpha(1+\alpha)}|D(j,r|y)| + \frac{(1-I(j))(1-I(r))}{(1+\alpha)}|D(j,r|\bar{y})|\right),$$

$$= \frac{1}{|I(r)|}\sum_j P(j)\left(\frac{(\alpha+I(j))(\alpha+I(r))}{\alpha(1+\alpha)}|D(j,r|y)| + \frac{(1-I(j))(1-I(r))}{(1+\alpha)}|D(j,r|\bar{y})|\right). \text{ (Proof of Formula (10))}$$

Since $-\alpha \leq I(j) \leq 1$, for any $a$, $b$ $\max_{I(j)}(a(\alpha + I(j)) + b(1-I(j))) = \max((\alpha+1)a, (\alpha+1)b)$. Therefore, let $a=\frac{\alpha+I(r)}{\alpha(1+\alpha)}D(j,r|y)$ and $b=\frac{1-I(r)}{\alpha}D(j,r|\bar{y})$, using (10) we have:

$$Dist(f(\mathbf{x},r), f(\mathbf{x},y)) \leq \frac{1}{|I(r)|}\sum_j P(j)\max\left(\frac{(\alpha+1)(\alpha+I(r))}{\alpha(1+\alpha)}|D(j,r|y)|, \frac{(\alpha+1)(1-I(r))}{(1+\alpha)}|D(j,r|\bar{y})|\right)$$

$$= \frac{1}{|I(r)|}\sum_j P(j)\max\left(\frac{(\alpha+I(r))}{\alpha}|D(j,r|y)|, (1-I(r))|D(j,r|\bar{y})|\right). \text{ (Proof for Formula (11))}$$

This theorem indicates that the distance between a RDE and a semi-perfect RDE in Formula (7) can be bounded by Inequity (10) or (11), which is determined by the following factors related to the reference feature, i.e., $I(r)$, $|D(j,r|y)|$ and $|D(j,r|\bar{y})|$. The theorem leads to the same result as Corollary 1 when $I(r) \neq 0$ and $D(j,r|y) = D(j,r|\bar{y}) = 0$ (conditionally independent). If $r$ and $j$ are not conditionally independent, a large $|I(r)|$, a small $|D(j,r|y)|$ and $|D(j,r|\bar{y})|$ can also lead to a small upper bound of distance. The bound in Formula (11) is looser than Formula (10), but there is no need to estimate each $I(j)$, which is usually difficult to learn accurately from limited labeled training data. Note that the estimation of $|D(j,r|y)|$ and $|D(j,r|\bar{y})|$ also requires the joint probabilities concerned with the class labels, which could be inaccurate due to the limited size of labeled data. Intuitively, there is connection between conditional independence and non-conditional independence, so we derive another bound including only the measure of non-conditional independence.

**Theorem 3:** *Given a RDE $f(\mathbf{x},r)$ and a semi-perfect RDE $f(\mathbf{x},y)$,*

*if $\forall j\ D(j,r|y)D(j,r|\bar{y}) \geq 0$, then:*

$$Dist(f(\mathbf{x},r), f(\mathbf{x},y)) \leq \frac{1}{|I(r)|}\sum_j P(j)\left|\frac{P(j,r)}{P(j)P(r)} - 1\right| + M \quad (12)$$

where $M = \frac{1}{sign(I(r))\alpha}\left(\sum_{\{j|D(j,r|y)\geq 0\}}P(j)I(j) - \sum_{\{j|D(j,r|y)<0\}}P(j)I(j)\right)$ (13)

**Proof:** combining Formula (10) and $D(j,r|y)D(j,r|\bar{y}) \geq 0$,

$$Dist(f(\mathbf{x},r), f(\mathbf{x},y)) \leq \frac{1}{|I(r)|}\sum_{j=1}^{m} P(j)\left(\frac{(\alpha+I(j))(\alpha+I(r))}{\alpha(1+\alpha)}|D(j,r|y)| + \frac{(1-I(j))(1-I(r))}{(1+\alpha)}|D(j,r|\bar{y})|\right)$$

$$= \frac{1}{|I(r)|}\sum_{\{j|D(j,r|y)\geq 0\}} P(j)\left(\frac{(\alpha+I(j))(\alpha+I(r))}{\alpha(1+\alpha)}\left(\frac{P(j,r|y)}{P(j|y)P(r|y)} - 1\right) + \frac{(1-I(j))(1-I(r))}{(1+\alpha)}\left(\frac{P(j,r|\bar{y})}{P(j|\bar{y})P(r|\bar{y})} - 1\right)\right) -$$

$$\frac{1}{|I(r)|}\sum_{\{j|D(j,r|y)<0\}} P(j)\left(\frac{(\alpha+I(j))(\alpha+I(r))}{\alpha(1+\alpha)}\left(\frac{P(j,r|y)}{P(j|y)P(r|y)} - 1\right) + \frac{(1-I(j))(1-I(r))}{(1+\alpha)}\left(\frac{P(j,r|\bar{y})}{P(j|\bar{y})P(r|\bar{y})} - 1\right)\right)$$

$$= \frac{1}{|I(r)|}\sum_{\{j|D(j,r|y)<0\}} P(j)\left(\frac{P(j,r)}{P(j)P(r)} - 1 - \frac{I(r)I(j)}{\alpha}\right) - \frac{1}{|I(r)|}\sum_{\{j|D(j,r|y)<0\}} P(j)\left(\frac{P(j,r)}{P(j)P(r)} - 1 - \frac{I(r)I(j)}{\alpha}\right)$$

$$\leq \frac{1}{|I(r)|}\sum_j P(j)\left|\frac{P(j,r)}{P(j)P(r)} - 1\right| + \frac{1}{sign(I(r))\alpha}\left(\sum_{\{j|D(j,r|y)\geq 0\}} P(j)I(j) - \sum_{\{j|D(j,r|y)<0\}} P(j)I(j)\right)$$

Here in order to make the following analysis easier we assume that the signs of $D(j,r|y)$ and $D(j,r|\bar{y})$ are equivalent, which is a much more relaxed assumption than that in Theorem 1. The bound avoids the estimation of $|D(j,r|y)|$ or $|D(j,r|\bar{y})|$, while it introduces a new term $M$. Using Formula (2) and (13), we can bound $M$ as follow:

$$-\sum_j P(j) \leq -\frac{1}{\alpha}\sum_j P(j)|I(j)| \leq M \leq \frac{1}{\alpha}\sum_j P(j)|I(j)| \leq \sum_j P(j) \qquad (14)$$

In the experiment we found $M$ changed slightly with different $r$ and was much smaller than the first part. In this case, we can use only the first part of Formula (12) for reference feature selection, which suggests selecting the reference feature with high precision in individual performance (a big $|I(r)|$) and high independence with other features (a small $\sum_j P(j)\left|\frac{P(j,r)}{P(j)P(r)} - 1\right|$).

## 2.2 Algorithm Design

The theory can be used for algorithm design in several aspects: 1) the bounds in Theorem 2 and 3 can be used for reference feature selection. 2) Even if the performance of individual RDE is inferior, combining multiple RDEs derived from different reference features may lead to a higher performance. 3) Prune the original features that yield a high distance. Note that different from Method 1, Method 3 is to select the original feature $j$ rather than the reference feature $r$. For example, if one of the original features is the same as or highly correlated to the reference feature, removing it from feature set **X** tends to lower the upper bounds in Theorem 2 and 3. In the following, we present a simple algorithm that addresses the strategies:

1) Rank candidate reference features by $\frac{1}{|I(r)|}\sum_j P(j)\left(\left|\frac{P(j,r)}{P(j)P(r)} - 1\right|\right)$ in ascending order and select top $k$ reference features.

2) Construct $k$ RDEs with the selected reference features in Step 1.

3) For each RDE remove any original feature $j$ with $\left|\frac{P(j,r)}{P(j)P(r)} - 1\right| > t$

4) Build a classifier using the decision score of each pruned RDE as a feature, and train the classifier with labeled examples.

Here the candidate reference features can be the existing features in labeled data, and $|I(r)|$ can be estimated from labeled data, and $\sum_j P(j)\left(\left|\frac{P(j,r)}{P(j)P(r)} - 1\right|\right)$ from unlabeled data. Since we assume that features are Boolean types, all the probabilities can be estimated based on the frequencies in labeled or unlabeled data:

$$P(A) = \frac{N_u(A)}{N_u}, \; P(A,B) = \frac{N_u(A,B)}{N_u}, \; P(A|B) = \frac{N_u(A,B)}{N_u(B)}, \; I(A) = \frac{N_l(A,y) - \alpha N_l(A,\bar{y})}{N_l(A)}$$

where $A$ and $B$ refer to any features, $N_u(A)$ is count of $A$ in unlabeled data, $N_u(A,B)$ is the co-occurrence count of $A$ and $B$ in unlabeled data, and $N_u$ is the total number of examples in unlabeled data. Similarly, $N_l$ means labeled data. Note that labeled data is used only for the estimation of $I(A)$.

In the algorithm, there are two parameters, i.e., $k$ (the number of reference feature) and $t$ (the threshold for feature selection) required to be selected in practice. The parameters can be tuned in the labeled training data and also depends on the specific classifier used in Step (4).

This algorithm can be viewed as an extension of our previous work feature coupling generalization (FCG) [8], a framework for generating new features from co-occurrence in unlabeled data. It achieved state-of-the-art performance in several challenging datasets. However, FCG is more of a general strategy rather than a specific algorithm, that is, there is a lot of heuristics in the design of FCG, and no theoretical analysis to justify why it works, which seriously limit its applications. The theoretical study here shed light on why FCG works in a special case where the feature co-occurrence measure is $P(r|j) - P(r)$, and the algorithm presented here is much easier to be applied to various real-world tasks.

## 3        Experiments

The dataset consists of 13 million MEDLINE abstracts. The task is to determine whether an article belongs to one of the 10 classes, the first 10 main headings in MeSH tree (see also Table 1). The reason we used the 10 binary text classification tasks is that the dataset has 13M labeled examples, so it is able to estimate a more accurate semi-perfect classifier than other smaller datasets, which is crucial in this work, and the MEDLINE data is clean and free available, so the experiment can be followed easily. Here we randomly selected 5,000 examples from the 13M ones as labeled data, other 5,000 examples for testing and the rest of them as unlabeled data. Sampling was done 5 times, and the mean and standard deviation of AUCs were reported in Table 1 for each task. AUC was used for evaluation because some classes are highly imbalanced where the AUC can be a more stable measure than F-score and Accuracy, and it avoids the impact of tuning thresholds. Two feature sets were examined in this work: unigrams and bigrams of words. Three types of RDEs were investigated: SuRDE is supervised RDE that used the label $y$ as reference feature (Case 1 in Section 1). SeRDE is the ensemble of semi-supervised RDEs (the algorithm in Section 2.2). PerfRDE is SuRDE trained with the 13M labeled examples, which was treated as an approximation of semi-perfect RDE. For SeRDE, $k$ was set at 100, $t$ was 20 and the candidate reference features were words with the count over 20 in each training set, since the estimation of $I(r)$ from 5k labeled examples could be inaccurate for low frequency words. Logistic regression in Weka [9] (with "ridge" parameter set at 40 for all the tasks) was used to integrate the RDE-based features in the SeRDE. Naïve Bayes classifier with Laplace smoothing and toolkits for large scale SVM [10] and Logistic regression [11] were used as baselines. Their parameters were tuned to be the optimal on the test set of each sampling.

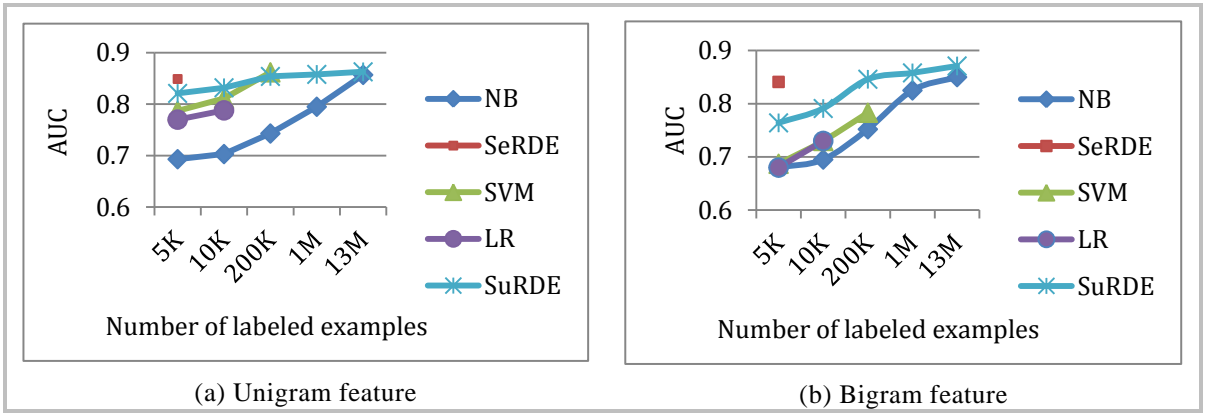

(a) Unigram feature                                      (b) Bigram feature

Figure 1: Average performance varied with the size of labeled training data

In Figure 1, we compare the average performance of classifiers trained on different sizes of labeled data, and it shows that given more labeled examples, the performances increase and the discrepancy between different classifiers become small. In these tasks, SVM and Logistic regression were difficult to work on more than 200K training examples due to the complexity of optimization algorithms. The best AUCs from RDEs on the two feature sets are 86% and 87.1% respectively, which are treated as the approximation of semi-perfect RDEs, the goals we aim at. It is encouraging to see that using 5,000 labeled examples, the performance of SeRDE (84.9% and 84.1%) is better than supervised methods just as moving towards the semi-perfect classifiers. A much larger improvement for bigrams than unigrams indicates that RDE can be especially effective in solving the data sparseness problem caused by bigrams. When the training data is small, unigrams perform better than bigrams, and vice versa, which implies that semi-supervised learning provides more opportunities for feature engineering, since some poor features in supervised learning can become good features using unlabeled data. Also it is notable that supervised RDE from 5,000 labeled examples performs much better any other supervised classifiers, while SVM performs best on 200K labeled examples on unigram features.

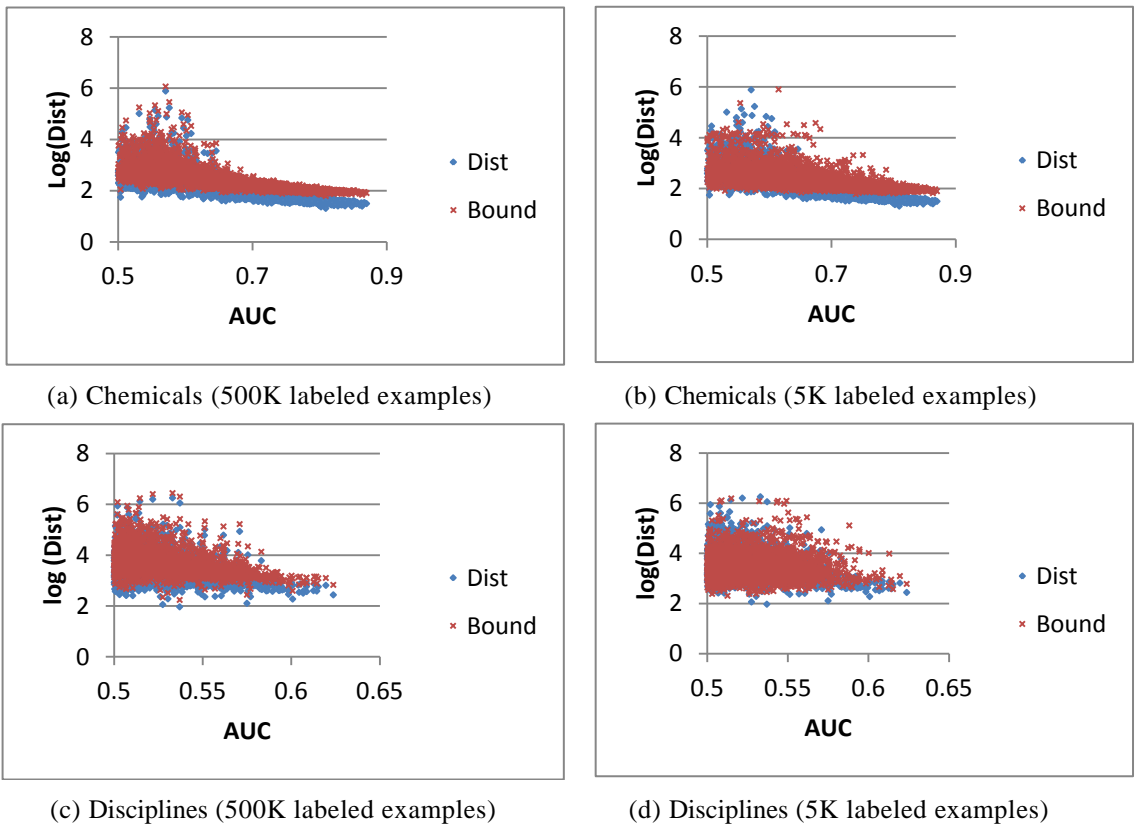

(a) Chemicals (500K labeled examples)  (b) Chemicals (5K labeled examples)

(c) Disciplines (500K labeled examples)  (d) Disciplines (5K labeled examples)

Figure 2: Relation between AUC of RDE, the distance (RDE vs. Semi-perfect RDE) and the bound of the distance.

Dist – the distance in Formula (7); Bound – the first part in Formula (12). The X axis is AUC and Y axis is the log of each distance based on 10. Each point relates to a RDE with a word-based reference feature. The size of labeled data used to estimate the bound is given in bracket. The best and worst performing classes ('Chemicals' and 'Disciplines') were selected for observation.

From Figure 2, we can see clearly that the AUC achieved by RDE is highly correlated to the distance as well as bound in the Theorem 3 especially when the AUC is high, which convinces our effort to improve the performance by minimizing the distance. In addition, Figure 2 shows that the bound and real distance have the similar trend, and the bound estimated from 5K labeled data and 500K unlabeled data have similar shapes to that from 500K labeled data. This demonstrates that the bound can be useful for algorithm design in practice, which was also convinced by the good performance of SeRDE. In addition, a surprising finding is that in some task, e.g., the "Chemicals" class, an RDE from only one word-based reference feature can lead to a high AUC over 85%, since every data point in Figure 2 is concerned with one reference feature. Therefore, in some applications, if we don't have labeled examples but know some good reference feature, for example, with high imbalance degree in Formula (2) based on our prior knowledge, a high classification performance can be also obtained via RDE without any labeled training data.

For each task we compare the performance of several semi-supervised learning algorithms including Co-training, TSVM and SeRDE in Table 1. Here the base classifier for co-training is supervised RDE, since it is the best supervised classifier in the experiment and scalable well. For co-training, we split the feature set into two views: the first half of the article and the rest half. We found that there was no significant change after co-training, and the improvement in 'Anthropology' was just because of the divide-and-conquer of the two views but not the introduction of unlabeled data. TSVM in SVM-light [5] was used in the experiment. It was also not able to improve the performance over supervised SVM. One possible reason is that the size of both labeled and unlabeled data are big in this work, while in most previous works these two methods were reported to be effective on very small size of data, e.g., several labeled examples and thousands of unlabeled ones [2]. It is also notable that in many of the tasks SeRDE outperformed SVM by 5% - 10%, and performed as well as the semi-perfect RDE, while SVM has been considered as one of the most effective classifiers for text classification. It also justified the strategy in Section 2.2 that the ensemble of RDEs can move closer to the semi-perfect RDE, although the performance of individual RDE may be inferior. We also tried TFIDF weighting scheme for SVM as well as other classifiers, but the results were not as good as binary features, so they were not reported in this paper.

Table 1: Comparison of AUC (%) on each task.

The size of labeled data used is in bracket. The features are unigram bag-of-words. There are 13M labeled examples used in "PerfRDE", and 5K used in the other methods. The best results from 5K labeled data were bolded.

| Task | SVM (5K) | SuRDE (5K) | Co-training (5K) | TSVM (5K) | SeRDE (5K) | PerfRDE (13M) |
|---|---|---|---|---|---|---|
| Anatomy | 82.6±0.4 | 82.3±0.6 | 82.2±0.6 | 81.6±0.4 | **85.8±0.8** | 84.9±0.5 |
| Organisms | 82.1±1.7 | 87.1±1.4 | 87.6±1.2 | 79.3±1.5 | **90.2±1.2** | 92.8±0.8 |
| Diseases | 83.7±0.7 | 85.1±0.5 | 85.9±0.2 | 80.3±0.8 | **87.7±0.5** | 87.9±0.2 |
| Chemicals | 87.6±0.4 | 88.6±0.2 | 88.2±0.3 | 86.1±0.5 | **91.8±0.2** | 91.2±0.2 |
| Analytical | 70.2±0.9 | 71.1±0.8 | 71±0.7 | 69.4±0.8 | **74.4±0.8** | 73.8±0.8 |
| Psychiatry | 83.8±0.4 | 88.9±0.6 | 89.1±0.8 | 81.6±0.9 | **91±1.0** | 91.8±0.5 |
| Phenomena | 80.1±0.2 | 80.4±0.3 | 80.6±0.5 | 78.8±0.5 | **82.7±0.5** | 82.6±0.2 |
| Disciplines | 65.8±0.7 | 70.2±1.5 | 70.5±2.3 | 63.6±1.0 | **74.2±1.6** | 78.9±1.6 |
| Anthropology | 76±0.5 | 84.1±0.4 | 85.7±0.3 | 74.6±0.6 | **86.3±0.7** | 88.3±0.8 |
| Technology | 74.7±1.7 | 82.8±0.8 | 80.8±2.2 | 72.5±1.9 | **85.4±1.1** | 88±1.5 |
| Average | 78.7±0.2 | 82.1±0.1 | 82.2±0.2 | 76.8±0.4 | **84.9±0.2** | 86±0.3 |

## 5    Conclusions

In this work, we found an interesting theory behind RDE, a simple classifier with pretty good performance both in supervised and semi-supervised settings. The theory as well as algorithm provides a powerful tool for exploiting the "big data". There are many directions that can be addressed in the future:

1) Applying it to various machine learning tasks with a lot of unlabeled data. For binary features, it can be used in a straightforward way as the experiment in this work. For real value features, one strategy is to use discretization technique or binary classifier to convert the features into binary ones, and another way is to give density estimation for the probabilities in RDE like generative models.

2) Developing better algorithms based on the theory, for example, learning a reference feature by optimizing the bounds, and jointly selecting original features and reference features, which will lead to many novel machine learning algorithms.

3) Exploiting potentially useful features for RDE using feature conjunction, such as the bigram features in the experiment, which were not good features in 5K labeled data but worked well in 13M data. It implies that the effectiveness of features even in the same task is not constant but varies with the size of both labeled and unlabeled data. For the tasks that have already reached the semi-perfect RDE (e.g., some tasks in Table 1), the only way to improve the performance further is to build another semi-perfect RDE with better performance using feature conjunction (e.g., high order n-grams), and find another set of reference features that fit the new feature set. Therefore, machine learning can be described as a process that we first establish a feature set that could potentially lead to a good semi-perfect classifier, and then find good reference features by learning from as much unlabeled data as possible so as to move close to this goal. There would be a lot of theoretical as well as experimental work to do in this new area.

## Acknowledgments

This work was supported in part by grants from "the Fundamental Research Funds for the Central Universities No. DUT11RC(3)85" and "China Postdoctoral Science Foundation funded project No. 2012M511147."